\documentstyle[aclap]{article}

\newcommand{\cosma}{{\sc Cosma}}
\newcommand{\Cosma}{{\sc Cosma}}

\newcommand{\email}{e-mail}
\newcommand{\emails}{e-mails}
\newcommand{\smes}{{\sf smes}}
\newcommand{\il}{{\sc Imas}}
\newcommand{\dito}{{\sc DiTo}}
\newcommand{\tsdb}{{\sf tsdb}}

\newlength{\itemlen} 

\newenvironment{desclist}%
{\begin{list}{}{
  \setlength{\labelwidth}{\itemlen}
  \setlength{\labelsep}{0em}
  \setlength{\leftmargin}{\itemlen}
  \setlength{\parsep}{1ex}
  \setlength{\listparindent}{0em}}}%
{\end{list}}

\newlength{\descilen} 
\newcommand{\ditem}[1]{\item[{\rm #1 \hfill}]}

\author{
{\bf Stephan Busemann, Thierry Declerck, Abdel Kader Diagne,}\\
{\bf Luca Dini, Judith Klein, Sven Schmeier}\\
DFKI GmbH \\ 
Stuhlsatzenhausweg 3, 66123 Saarbr\"{u}cken, Germany\\
{\tt busemann@dfki.uni-sb.de}
}
\title{\vspace{-0.5in}Natural Language Dialogue Service \\ for 
Appointment Scheduling Agents\thanks{This work has been supported by
a grant from the German Federal Ministry of Education, Science, Research and
Technology (FKZ~ITW-9402).}}
\date{}

\begin{document}
\include{epsf}
\bibliographystyle{acl}
\maketitle
\vspace{-0.5in}
\begin{abstract}
Appointment scheduling is a problem faced daily by many individuals and
organizations. Cooperating agent systems have been developed to
partially automate this task. In order to extend the circle of 
participants as far as possible we advocate the use of natural
language transmitted by \email. We describe \cosma,  a fully
implemented German language server for existing appointment 
scheduling agent systems. \Cosma\ can cope with multiple dialogues 
in parallel, and accounts for differences in dialogue 
behaviour between human and machine agents. NL coverage of the
sublanguage is achieved through both corpus-based grammar development 
and the use of message extraction 
techniques.
\end{abstract}

\section{Motivation}

Appointment scheduling is a problem faced daily by many individuals and
organizations, and typically solved using communication in natural
language (NL) by phone, fax or by mail. In general,  cooperative
interaction between several participants is required.
Since appointments are often scheduled only after a sequence of
point-to-point connections this will, at times, necessitate repeated
rounds of communication until all participants agree to some date and
place. This is a very time-consuming task that should be automated.  

Systems available on the market allow for calendar and contact management.  
As \cite{Bus:Mer:95} point out in a market survey, all planning and
scheduling activity remains with the user. 
Cooperative agent systems developed in the field of Distributed AI are
designed to account for the scheduling tasks. Using distributed rather than
centralized calendar systems, they not only guarantee a maximum
privacy of calendar information but also offer their services to members
or employees in external organizations. Although agent systems allow 
users to automate their scheduling tasks to a 
considerable degree, the circle of participants remains restricted to users
with compatible systems. 

To overcome this drawback we have designed and implemented \cosma,
a novel kind of NL  dialogue systems that
serves as a German language front-end system to scheduling agents. 
Human language makes agent services available to a much broader public.
\cosma\ allows human and machine agents to participate in appointment
scheduling dialogues via \email. We are concerned with meetings all 
participants should attend and the date of which is negotiable.

\section{Design guidelines}

\cosma\ is organized as a client/server architecture. The server
offers NL dialogue service to multiple client agent systems. 
Up to now, three different types of agent systems have been hooked up
to the NL server. Agents developed in-house 
were used for the early system described in \cite{Bus:Oep:Hin:94}. In 
a subsequent version, the MEKKA agents developed by Siemens~AG 
\cite{Lux:Bom:Ste:92} have been adapted. We present in Section~\ref{pasha} 
a third kind of client system, the PASHA~II user agent.

Given the use of distributed calendar systems, techniques used by both
human and machine agents for cooperatively scheduling appointments 
must be based on negotiation dialogues. However,
human dialogue behaviour differs from interaction between machine
agents considerably, as will be discussed in Section~\ref{pasha}. 
A human-machine interface to existing appointment
scheduling agent systems should  comply to the following requirements:
\begin{itemize}
\item Human utterances must be analyzed to correspond closely to agent actions.
\item Machine utterances must conform to human dialogue strategies.
\end{itemize}
Artificial communication languages have been designed for human
discourse, e.g.\ \cite{Sidner:94}, as well as for agent-agent interaction,
e.g.\ \cite{Ste:Bur:Kol:93}. What would be needed for \cosma\ is a mapping
between strategies implemented in such languages. Since the type of
agent system connected to the \cosma\ server is not restricted by its
dialogue behaviour, preference was given to implement application-dependent 
mappings instead of developing a generic formalism. 
As a consequence, \cosma\ operates with general and reusable processing
modules that interpret domain- and task-specific data. 

The same principle was also adopted for NL analysis.
The server must analyze human-generated text and verbalize
machine-initiated goals. For a plausible application, the server must be:
\begin{itemize}
\item complete with respect to a sublanguage: all relevant information related
to appointments must be analyzed,
\item sufficiently robust to deal with inconsistent analysis results.
\end{itemize}

Within the HPSG-based approach to grammar description 
adopted for the early system \cite{Usz:Bac:Bus:94}, achieving 
these goals turned out to be
difficult. This ``deep'' approach to NLU describes NL 
expressions at general linguistic levels (syntax and surface
semantics), and attempts to capture the complete meanings of all and
only the grammatical sentences.  However, an NL system in a 
realistic application should not fail on unexpected input. Moreover,
the surface semantic representations derived by the grammar were too close
to NL for an agent system to deal with.

With the present version of the NL server
these problems are solved by adopting a ``shallow'' analysis approach,
which extracts meanings
from those portions of a text that are defined as interesting and
represents them in an agent-oriented way.
Instead of failing on unexpected input,
shallow parsing methods  always yield
results, although they may not capture all of the meaning intended by 
the user. By just describing the verbalizations
of relevant information, shallow parsing grammars are highly
domain-specific and task-oriented. 
In \cosma, shallow analysis is divided up into an application of the
message extraction component \smes\ (discussed in Section~\ref{smes})
and a semantic analysis component \il\ (Section~\ref{sem}). The
former extracts appointment-related information from users' input
texts. It is based on finite-state 
automata that were defined with help of an annotated corpus of \email\ 
messages. The task of the latter is to derive a client-oriented
semantic representation, including the communicative intention
and the complete specification of 
time points needed, which is based on context and semantic inferences.

The robustness requirement is fulfilled by recognizing failures  within
the server during semantic analysis, and possibly within the client systems, 
and by clarification dialogues (cf. Section~\ref{failure-handling}).
\begin{figure*}
\hspace*{5mm}
\epsfbox{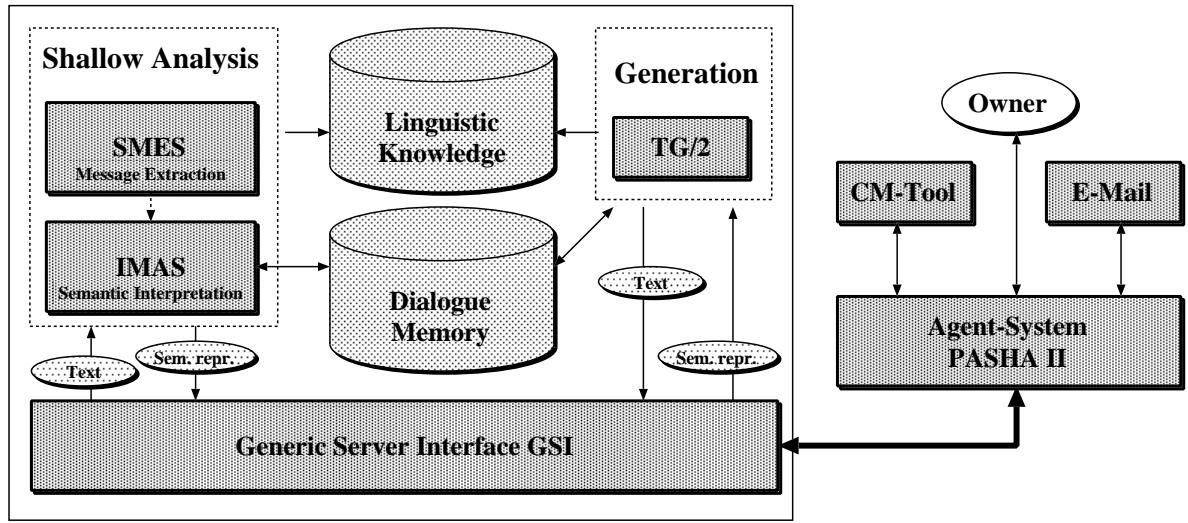}
\caption{The \protect\cosma\ architecture: a client connected to a
server instance may issue requests to receive a
semantic representation for a text, or to generate
a text from a semantic representation. The generic server interface
invokes the necessary server
processes and maintains interaction with the client.}
\label{nls-arc}
\end{figure*}

After an overview of generation in \cosma\ (Section~\ref{gen}) we discuss 
component interaction in Section~\ref{arch}. A novel type
of object-oriented architecture is needed to treat multiple
dialogues in parallel. Virtual partial system instances are maintained
as long as a dialogue is going on. 
One such instance is shown in Figure~\ref{nls-arc}.

\section{A complete sample dialogue}
A complete sample dialogue taken from the system's
present performance will serve as a reference throughout the paper.
Every utterance is numbered and labeled; the labels indicate speakers.
We assume a three-party \email\ negotiation between a human (H), who does not
use a scheduling agent system, and two machine agents (A, B) that
schedule appointments for their respective owners. In the server,
human interactions with multiple 
machine partners are treated  as different
NL dialogues (in the present case between H and A, and H and B). 
In what follows, H is the 
initiator, but \Cosma\ also copes with machine-initiated
dialogues (cf.\ Section~\ref{pasha}).\footnote{Agent-agent
interaction is based on a formal representation language rather than on
NL.}

\fussy
\begin{desclist}
\ditem{\parbox[t]{\itemlen}{(01)\\H}}
\parbox[t]{\descilen}{Ich 
w\"{u}rde Sie gern am Montag, dem 2.\ 11.\ 96 wegen der
bevorstehenden Projektbegutachtung treffen. [I would like to meet you
on Monday Nov.\ 2 1996 about the upcoming project
review.]}
\ditem{\parbox[t]{\itemlen}{(02)\\A, B}}
\parbox[t]{\descilen}{COSMA 
hat die folgende Zeitangabe    
verstanden, die nicht konsistent ist: Montag, den 2.\ 11.\ 1996.
K\"{o}nnten Sie bitte den Wochentag oder das Datum korrigieren?
[COSMA has understood the following time expression, which is not 
consistent: Monday, Nov.\ 2 1996.
Could you please correct the weekday or the date?]}
\ditem{\parbox[t]{\itemlen}{(03)\\H}}\parbox[t]{\descilen}{Ich meinte
nat\"{u}rlich Montag den 4.\ 11.
[I meant of course Monday, Nov.\ 4.]}
\ditem{\parbox[t]{\itemlen}{(04)\\A}}\parbox[t]{\descilen}{Am 4.\ 11.\ 1996 
pa\ss t es bei mir zwischen 13 und 18 Uhr.
[Nov.\ 4, 1996 would suit me between 1  and 6 p.m.]}
\ditem{\parbox[t]{\itemlen}{(05)\\B}}\parbox[t]{\descilen}{Leider kann ich am
4.\ 11.\ 1996 nicht kommen.
[Unfortunately I can't come on Nov.\ 4 1996.]}
\ditem{\parbox[t]{\itemlen}{(06)\\H}}\parbox[t]{\descilen}{Der vorige 
Vorschlag wurde abgelehnt. K\"{o}nnen wir uns stattdessen am Dienstag 
tref\-fen?
[The previous proposal was rejected. Can we meet instead on 
Tuesday?]}
\ditem{\parbox[t]{\itemlen}{(07)\\A}}\parbox[t]{\descilen}{Zu folgenden 
Zeiten geht es bei mir: am 5.\ 11.\ 1996 zwischen 8 und 12 Uhr 
und zwischen 16 und 18 Uhr.
[The following times would suit me: Nov.\ 5, 1996 between 8 
and 12 a.m.\ and between 4  and 6 p.m.]}
\ditem{\parbox[t]{\itemlen}{(08)\\B}}\parbox[t]{\descilen}{Am 5.\ 11.\ 1996 
pa\ss t es bei mir zwischen 10 und 12 Uhr.
[Nov.\ 4, 1996 would suit me between 10  and 12 a.m.]}
\ditem{\parbox[t]{\itemlen}{(09)\\H}}\parbox[t]{\descilen}{Wie w\"{a}re 
es dann um 10? [How about 10 then?]}
\ditem{\parbox[t]{\itemlen}{(10)\\A, B}}\parbox[t]{\descilen}{Ich sage den 
5.\ 11.\ 1996 um 10 Uhr zu.
[I accept Nov.\ 5, 1996 at 10 a.m.]}
\ditem{\parbox[t]{\itemlen}{(11)\\H}}\parbox[t]{\descilen}{Das Treffen 
wird also am 5.\ 11.\ 1996 um 10 Uhr stattfinden. 
[The meeting will take place on Nov.\ 5, 1996 at 10 a.m.]}
\end{desclist}\sloppy
In (01) H sends by mistake an inconsistent temporal expression to A and
B, giving rise to clarification dialogues initiated by each of A and B
(02). The repair provided 
by H (03) is underspecified with respect to clock time (see also (06)), 
hence the agents offer free time slots in accordance to their calendars
(04, 07, 08). These time slots are blocked until further
notice.\footnote{Cancellations of reserved slots due to a high-priority
request are a straight-forward extension of the present coverage.}
Since B rejects the proposed date (05), a new loop is started 
by H (06). When H notices that Tuesday is promising, she chooses to
refine her proposal by suggesting a clock time (09).
Dynamic context knowledge allows the server to reconstruct a full time
specification that is interpreted by the agents as an alternative proposal.
Refinements can thus be dealt with completely in the server, whereas the agents
may or may not have a concept of refinement. After all agents accept a
proposal, the date is confirmed by the initiator (11). Upon receipt of
the confirmation, the agents fix the date in their calendars. Server
and agents consider the dialogues as completed.

\section{Dialoging scheduling agents}
\label{pasha}
\subsection{The PASHA~II system}
PASHA~II agents \cite{Sch:Sch:96} are designed according to the InterRaP 
agent architecture \cite{Fischer+95a}, a layer-based agent model that 
combines deliberative and reactive behaviour. The ``heart'' of an agent 
is the {\em cooperative planning layer}, in which
negotiation strategies are represented as programs and executed by a
language interpreter. This supports easy modification and exchange of
plans. The {\em local planning layer\/} consists of a constraint planner which
reasons about time slots in the agent's (i.e. its owner's) calendar. 
In contrast to the planning layers, the {\em behaviour-based layer\/}
consists of the agent's basic reactive behaviour and its procedural
knowledge. The {\em world interface\/} realizes the agent's sensing and acting
capabilities as well as the connection to its owner. PASHA~II 
agents are connected to the Unix {\sc cm} calendar management tool,
but can easily be hooked up to other calendar systems. 

PASHA~II agents
are easily adapted to the owner's preferences. For instance,
any time slots the owner does not wish the agent to use can be
blocked. By virtue of this mechanism, a working day could be defined
as an interval from e.g.\ 8 a.m.\ until 6 p.m.\ except for Saturdays, Sundays 
and holidays. Moreover, gaps between appointments may be specified
in order to permit sufficient time between meetings. 

\subsection{Adapting agents to the \protect\cosma\ server}
Taking PASHA~II as a representative, we describe the requirements 
for an agent system to connect to the \cosma\ server.


{\bf Interface to the server.} The four main modules include
the basic TCP/IP connection to the server; 
a parser of semantic representations of the server's analysis results, which
yields PASHA~II structures;
an instantiation mechanism for semantic generation templates; and
a control regime that keeps track of the
current dialogue. The control regime 
confirms results of the server, or it activates 
the server's backtrack mechanism
if the semantic representation received does not fit within the 
current dialogue step, or it issues a request for repair if 
backtracking should not yield any further results.

{\bf Receiving and sending \email.}
The PASHA~II interaction mechanism includes, besides
communication via TCP/IP protocols,  \email\ interaction.
The agent may poll its owner's mailbox or have one of its own. 
Either the agent or its owner is referred to as actor in 
the agent's \email\ messages (see Section~\ref{gen}).

{\bf Dialogue behaviour.}
An agent has to generate and
understand different dialogue actions represented by corresponding
{\em cooperation primitives} such as proposing, accepting, rejecting,
canceling or fixing a meeting \cite{Ste:Bur:Kol:93}.  

Agent-agent interaction usually relies on an initiating agent being
responsible for the success of a negotiation. The initiator's broadcast
proposal is triggered by its owner, who determines partners, duration and
an interval within which the appointment should be scheduled. The
agent proposes the first slot in the interval that is available
according to its calendar. In case of a
rejection of one or more participants, the initiator would
continue to propose new time slots to all partners until everyone agrees 
to a common date or there is no such slot within the interval.
Note that in case of rejection (see (05)) 
PASHA~II agents do not use counter-suggestions.

In human-human negotiation, efficiency is a major goal. Humans often follow
the {\em least effort principle\/} \cite{Dahlbaeck:92}: 
the initiator broadcasts a proposal 
including a time interval within which the meeting should take place
(e.g.\ (03)) and expects refinements or counter-proposals from the 
participants. As the example shows this may imply the use of
underspecified temporal descriptions.
This strategy requires less communication
because a greater amount of information is exchanged in one dialogue step
between the participants.

Handling  underspecified temporal information by offering free time slots (see
(04), (07), and (08))
is among the extensions of PASHA~II at the local planning layer. Note that this
strategy can be instantiated in different ways, as becomes clear from
dealing with expression such as {\em next week}:
Only a selection of free time slots can be provided here, which is
explicitly marked using e.g.\ {\em for instance}.
Moreover, we consider it indispensable to have agents understand and generate
counter-proposals to avoid inefficient plain rejections like (05).

\section{Covering the domain language}
\label{smes}

\subsection{Corpus-based annotation}

In order to determine the coverage of the sub-language relevant for the
application  and to measure progress during system development, a
corpus of 160 \emails\ was selected as reference material from several
hundred \emails\ collected from the domain of appointment scheduling. 
The \emails\ were manually analyzed and annotated with major syntactic
and semantic features as well as speechact information. A combination of two
relational database systems was employed to ease the storage,
maintenance, extension and retrieval of the NL
data: \\
(i) \dito\ \cite{Ner:93}, a full text database where the 
\emails\ can be accessed, \\
(ii) \tsdb\ \cite{Oep:95}, an elaborated fact database which permits the
extraction of specific linguistic constructions together with the
associated linguistic annotations.\footnote{\dito\ and \tsdb\ entries are linked
via \email\ identifiers.}

\begin{figure}[ht]
\begin{center} 
\begin{small}
\begin{tabular}{|p{2.1cm}|p{4.9cm}|} \hline
Annotation      &  Example \\ \hline \hline
\multicolumn{2}{|l|}{{\bf Prepositional Phrases:} Wie w\"are es [How about] ...} \\ \hline
PP\_temp        &  {\it in dieser Woche?}  [{\em in this week?}] \\
PP\_temp-date   &  {\it am 4.11?} [{\em on the 4th of Nov.?}] \\
PP\_temp-day    &  {\it am Montag?}  [{\em on Monday?}] \\ 
PP\_temp-dur    &  {\it von 8 bis 12?}  [{\em from 8 to 12?}] \\
PP\_temp-time   &  {\it um 10?}  [{\em at 10?}] \\ \hline
\multicolumn{2}{|l|}{{\bf Noun Phrases: } Ich komme [I come] ...}  \\ \hline
NP\_temp       &   {\it zwei Stunden sp\"ater. } \\
               &   [{\em two hours later.}]  \\
NP\_temp-date  &   am Montag, {\it den 4. 11.} \\
               &  [on Monday, {\em the 4th of Nov.}] \\
NP\_temp-day   &   {\it Montag}, 14 h. [{\em Monday}, 2 pm.]\\
NP\_temp-time  &  Montag, {\it 14 h.}  [Monday, {\em 2 pm.}] \\ \hline
\end{tabular}
\end{small}
\end{center}
\caption{Semantic annotation of PPs and NPs (annotated linguistic 
material in {\em italics})}     
\label{annotations}
\end{figure}

The annotation work is based on the TSNLP framework \cite{Leh:96}
where detailed category and function lists are defined for the structural and
dependency structure annotation of linguistic material for NLP test
suites. 
For \cosma, the 
classification has been extended according to semantic information
relevant for the appointment domain. For instance, PPs and NPs
were specified further, introducing a more fine-grained
semantic annotation for temporal expressions, as is shown in
Figure~\ref{annotations}.

The results of database queries provided valuable insights into the
range of linguistic phenomena the parsing system must cope with in the
domain at hand. Grammar development is guided by a frequency-based priority
scheme: The most important area -- temporal expressions of
various categories -- followed by basic phenomena including different
verbal subcategorizations, local and thematic PPs, and the verbal
complex are successfully covered. 


\subsection{Message extraction with \smes }

The message extraction system \smes\ \cite{Neu:Bac:Bau:97} is a core engine
for shallow processing with a highly modular architecture. Given an
{\tt ASCII} text, \smes\ currently produces predicate argument structures
containing shallow semantic analyses of PPs and NPs. 
The core of the system consists of:
\begin{itemize}

\item a tokenizer, which scans the input using a set of regular
expressions to identify the fragment patterns (e.g. words,
date expressions, etc.),

\item a fast lexical and morphological processing of 1,5 million
German word forms,

\item a shallow parsing module based on a set of finite state
transducers,

\item a result combination and output presentation component.
\end{itemize}
Based on the information delivered by the morphological analysis
of the identified fragment patterns, the system performs a constituent
analysis. In order to combine complements and adjuncts into predicate-argument
structures, special automata for verbs are then activated over the
sequence of constituents analyzed so far. Starting from the main 
verb\footnote{If no verb is found, a ``dummy'' entry triggers 
processing of verbless expressions, which occur frequently
in \email\ communication.}, a
bidirectional search is performed whose domain is restricted by
special clause markers. \smes\ output yields information
about the utterance relevant for the subsequent semantic analysis.

\subsection{Semi-automatic grammar development}
\label{semi-automatic}


The concrete realization of the automata is
based on the linguistic annotations of the \email\ fragments in the corpus.
The annotations render a semi-automatic description of
automata possible. For instance, verb classification directly leads to
the lexical assignment of a corresponding automaton in \smes.
By deriving parts of the grammar directly from corpus annotations, 
maintenance and extension of the grammars are eased considerably.

On the other hand, corpus extension can be supported by \smes\
analyses. Existing automata can be used to annotate new material 
with available linguistic
information. Manual checking of the results reveals gaps in the
coverage and leads to further refinement and extension of the automata
by the grammar writer.

This way, grammar development can be achieved in subsequent feedback cycles
between the annotated corpus and \smes\ automata. The 
implementation of the annotation procedure based on the 
\smes\ output format is underway.

\section{Semantic interpretation}
\label{sem}
Semantic representations produced by \smes\ are mapped into a format
suitable for  the PASHA-II client by the \il\ component (Information
extraction Module for Appointment Scheduling). \il\ is based on a
domain-dependent view of semantic interpretation: information-gathering
rules explore the input structure in order to collect all and only the relevant
information; the resulting pieces of information are combined and
enriched in a monotonic, non-compositional way, thereby obtaining an IL
(Interface Level) expression, which can be interpreted by the agent systems. 
In spite of the non-compositionality of this 
process, the resulting expressions have a clear model-theoretic
interpretation and could be used by any system accepting first order
logic representations as input. 

IL expressions have been designed with the goal of
representing both a domain action that is easily mapped onto an agent system's
cooperation primitive, and the associated
temporal information, which should be fully specified due to contextual
knowledge. Temporal information is partitioned into {\tt
RANGE}, {\tt APPOINTMENT} and {\tt DURATION} information. {\tt RANGE}
denotes the interval {\em within which} a certain appointment has to
take place (e.g.\ in (03)). {\tt APPOINTMENT} denotes the
interval of the appointment proper (e.g.\ in (10)). 
Intervals in general are represented by their boundaries.
{\tt DURATION}, on the contrary, encodes the duration of
the appointment expressed in minutes. The backbone of an IL expression
is thus the following: 
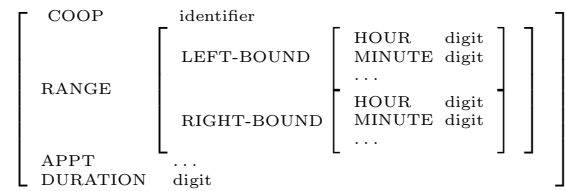
\vspace{1em}
\begin{tiny}
$ \left[ \begin{array}{ll} 
$ COOP$ & $ identifier$ \\
$RANGE$ & \hspace{-0.3cm} 
	\left[ \begin{array}{ll}
	$LEFT-BOUND$ & \hspace{-0.3cm} 	
		\left[ \begin{array}{ll}	
		$HOUR$ & \hspace{-0.2cm} $digit$ \\
		$MINUTE$ &\hspace{-0.2cm}  $digit$ \\
		\ldots & $ $
		\end{array}\right]\\ 
	$RIGHT-BOUND$ &	\hspace{-0.3cm} 
		\left[ \begin{array}{ll}	
		$HOUR$ & \hspace{-0.2cm} $digit$ \\
		$MINUTE$ & \hspace{-0.2cm} $digit$ \\
		\ldots &  $ $
		\end{array}\right] 
	\end{array}\right] \\
$APPT$ & \ldots \\
$DURATION$ & $digit$	
\end{array}\right]$
\end{tiny} 
\vspace{1em}

\il\ relies on three basic data structures. The {\bf sentence structure}
contains all the IL expressions obtained from the analysis of 
a single sentence. They are ranked according to their informativeness.

The {\bf text structure}
contains all the sentence structures obtained from the analysis of a 
whole message. Here ranking depends
not only on informativeness but also on ``dialogue
expectation'': sentence structures are favoured that contain a
domain action compatible with the IL expression previously
stored in the discourse memory.  As a result, the NL server will pass
to the client the most informative IL expression of the most
informative {\em and}  contextually most relevant sentence of the
analyzed text.\footnote{If the client is not satisfied with such an expression,
backtracking will pass the next-best structure etc.}

The {\bf discourse memory} is structured as a sequence containing  
all information collected during the dialogue. Thus it contains both IL
expressions  committed by the client and semantic input structures 
from generation. The discourse memory is used by  \il\ as a stack.---

The procedural core of  \il\ is represented by the transformation
of the input \smes\ representation  into a  set
of IL expressions. This process is organized into three steps: 

{\bf Linguistic extraction.} The semantic representation  of the input
\smes\ structure is explored by a set of rules in such a way that all
information relevant for the appointment domain is captured. For
every type of information (e.g.\ domain action,
hour of appointment, duration, etc.) a different set of
rules is used. 
The rules are coded in a transparent
and declarative  language that allows for a (possibly underspecified)
description of the \smes\ input (represented as a feature structure)
with its associated ``information gathering'' action. 

{\bf Anchoring.} Most  utterances concerning the domain of appointment
scheduling are incomplete at least in two respects. Either they  contain
expressions which need to be 
delimited in order to be pragmatically plausible (underspecification,
e.g.\ (09)), or they refer to
intervals  which are not explicitly mentioned in the sentence
(temporal anaphora). The first class includes probably any NL
time expression; even a simple expression such as  (01)
requires  some extralinguistic knowledge to be understood
in its proper contextual meaning (in (01) the
``working day'' interval of the respective day must be known).
The reconstruction of underspecified temporal
expressions  is performed by a set of
template filling functions which make use of  parameters specified by
the client system at the beginning of the dialogue.

Temporal anaphora include expressions such as {\em on Monday}, {\em
tomorrow}, {\em next month}, whose interpretation depends on the
discourse context.  Solving anaphoric and deictic relations 
involves a rather complex machinery which borrows many concepts from
Discourse Representation Theory. In particular, we assume a procedure 
according to 
which the antecedent of an anaphoric temporal expression is first
looked up in the IL expressions of the text already parsed (with a
preference for the most recent expressions); if no one is
found, the discourse memory is consulted to retrieve from 
previous parts of the dialogue a temporal expression satisfying the
constraints under analysis. If the search fails again, the
expression is interpreted deictically, and resolved w.r.t. to the time
the message  was sent.

{\bf Inferences.} IL expressions can be enriched and disambiguated by
performing certain inferences involving temporal reasoning. Besides
trivial cases of temporal constraint resolution, such as guessing the
endpoint of an appointment from its startpoint and its duration, our
inference engine performs disambiguation of domain actions by
comparing intervals referred to by different dialogue utterances. 
For instance, if an utterance $u$ describing an interval $I$ is ambiguous
between a refinement and a modification  and the previous utterance
refers to an interval $J$ including $I$, then $u$ can
be disambiguated safely as denoting a refinement. Analogous inferences
are drawn by just checking the possible combinations of domain actions
across the current dialogue (a rejection can hardly be followed
by another cancellation, a fixing cannot occur after a rejection,
etc.). 
The constraints guiding this disambiguation procedure 
are encoded as filters on the output of \il\ and reduce the set of 
pragmatically adequate IL expressions.

\subsection{Handling of analysis failures} 
\label{failure-handling}

Sometimes \il\ produces an output which cannot be used
by the PASHA-II client. This happens when the human message
is either too vague ({\em What about a meeting?}), or contains
an inconsistent temporal specification (as in (01)).   
In these cases \il\ stores the available information, and  the server
generates a request for clarification in order to recover the
necessary temporal specifications or to fix the already available
ones.  
This request is mailed to the human partner. It includes the
list of misspelled words found in the input message, which
may give the partner a clue for understanding the source of the error.
Once a clarification is provided, the server attempts to build an 
IL expression by merging and/or
replacing the information already available with the
newly extracted one (cf. (03)). 
If the resulting IL expression satisfies the
constraints on well-formedness, it is shipped to the PASHA-II client.
Otherwise the clarification subdialogue goes on along the same 
lines.

\section{Generation}
\label{gen}
Client systems usually want to express in NL a cooperation primitive 
and a date expression.  Hence NL generation is based on a semantic 
template filled by the client. Depending on its content the template is
unified with a prefabricated structure specifying linguistic-oriented
input to the generator. The same holds for failure messages, such as
(02), and for specifications of free time slots, as in (07), where
simple rules of aggregation take care not to repeat the full date
specification for each clock time mentioned.

The production system TG/2 \cite{Busemann:96b} proved to be sufficiently
flexible to accomplish this
task by its ability to generate preferred formulations first.
For instance, \cosma\ clients can parameterize TG/2 so as to 
refer to their owner by a first person pronoun or by a full name, 
or to use formal or informal form of addressing the human
hearer, or to prefer deictic time descriptions over
anaphorical ones.


\section{A novel architecture}
\label{arch}

A NLP server which can both provide a range of natural language services and 
process multiple dialogues for a variety of applications in parallel
requires (1) an architecture that ensures a high degre of 
reusability of NLP resources, (2) the availability of a robust interface  
that guarantees transparency and flexibility with respect to data 
representation and task specification, (3) client-driven server
parametrization, (4) support for incremental, distributed and
asynchronous robust data processing, and (5) advanced concepts for
synchronization with respect to parallel dialogue processing for
multiple clients.   
Due to the limited functionality of common architectural
styles \cite{garlan:93} with respect to these requirements, a novel
object-oriented, manager-based and generic architecture has been
designed and implemented. It combines techniques from different areas --
in particular, from object technology \cite{booch:94} and from
coordination theory including workflow management \cite{malone:91} --
and is based on two main concepts:
the cooperating managers approach ({\sc coconuts}) and the virtual system
architecture model.

\subsection{A manager-based approach} 
Managers in the {\sc coconuts} model are control units which coordinate or 
perform specific activities and cooperate with each other in a 
client/server form. Their responsabilities, properties, behaviour and interface
are determined by the classes they belong to. The prominent {\sc
coconuts} managers are: the data manager, which provides services
related to representation, printing, conversion and transmission of data;
the report manager, which supports specification, generation and
printing of processing reports; the global interface manager, 
which provides a generic server interface; the computing components  
managers ({\sc ccm}s), which
encapsulates the system's components and let them appear as servers;
and, finally, the workflow manager, which is the main control unit. 

\subsection{Coordination and control}
Coordinating internal system activities with respect to parallel dialogue
processing (including backtracking and failure recovery facilities)
requires very powerful and flexible mechanisms for task scheduling,
synchronization and control. In {\sc coconuts} this task is carried out
by the workflow manager, which also manages interdependencies between
these activities while avoiding redundant ones and controlling the flow
of work among the involved  managers (e.g., passing subtasks from one
manager to another in a correct sequence, ensuring that all fulfill
their required  contributions and taking default actions when
necessary). The behaviour and function of the workflow manager are 
determined by the following sequence of operations:  
 identifying and formulating a workflow goal, decomposing it into subgoals, 
 determining and allocating resources for achieving the subgoals, 
 elaborating and, eventually, executing an operation plan.  
It also provides a range of specialized exception handlers to ensure
robustness (see Section~\ref{failure-handling}).

\subsection{A generic server interface}
Flexible and reliable client/server communication is made possible by
the generic server interface module {\sc gsi}. It includes
a declarative, feature-based representation and task specification 
language {\sc ccl} and  an object-oriented communication and data
transfer module {\sc cci}. For {\sc ccl} a parser, a printer and an inference 
engine are available. {\sc cci} contains various kinds of 
{\it interface objects} containing higher-level protocols and
methods for reliable TCP/IP-based communication, data encoding/decoding and
buffering, as well as priority and reference management. Note that 
interface objects are accessible through their TCP/IP-based internet
addresses and can be associated to any component (cf.\
Figure~\ref{nls-arc}). This way, subsystems can, on demand, be used as servers,
e.g.\ \smes\ or the generator. 

\subsection{Integrating heterogenous components} 
Each {\sc Cosma} server component is encapsulated by a {\sc ccm}
(computing component manager), which makes its functionality available to
other managers. A {\sc ccm} has, among other things, a working (short-term)
memory, a long-term memory and a variety of buffers for storing and managing
computed solutions for subsequent use. Using these features a {\sc
ccm} easily simulates incrementality and realizes intelligent
backtracking by providing the computed solutions in a selective 
manner. A component can be released by a {\sc ccm} it is bound to when
the latter does no longer need its services; e.g.\ if the component
has already computed all solutions. This permits efficient resource
sharing, as several {\sc ccm}s can be associated to one component. Thus,
associating interface objects with {\sc ccm}s provides a flexible way
of realizing distributed processing performed by components implemented
in different languages and running on different machines. 
 
\subsection{The virtual system architecture}
The virtual system architecture allows for efficient
parallel dialogue processing. It is based on the concept of cooperating
object-oriented managers with the ability to define
one-to-many relationships between components and {\sc ccm}s.
The key idea consists in adopting a manager-based/object-based view
of the architecture shown in Figure~\ref{nls-arc}. This architecture
represents a virtual system (also called operation context), which is a highly
complex object consisting of a variety of interacting managers. It may
inherit from different classes of operation contexts, whose definitions are
determined by the underlying domains of application. Thus, multiple
dialogues are processed in parallel just by running each dialogue in a 
separate virtual system. As soon as a dialogue is completed, the assigned 
virtual system can be reused to process another one. Conceptually,
no constraints are made on the number of active virtual systems in the
server software. In order to ensure correct processing, a manager may
operate in only one virtual system  at a time.
Note that managers can still be shared by virtual systems and they behaviour
can vary from one system to another.

\section{Conclusion}
\label{concl}
We described \cosma, a NL server system for existing  machine agents
in the domain of appointment scheduling. The server is implemented in
Common Lisp and C. The PASHA~II agent is implemented in DFKI-Oz
\cite{Smolka:95}. 

Robust analysis of human \email\ messages is achieved
through message extraction techniques, corpus-based grammar
development, and client-oriented semantic processing and representation.
The virtual server architecture is a basis for the flexible use of
heterogeneous NLP systems in real-world applications including,
and going beyond, \cosma.

Future work includes extensive in-house tests that will provide
valuable feedback about the performance of the system. 
Further development of \cosma\ into an industrial prototype is envisaged.

\end{document}